\documentclass[runningheads]{llncs}
\usepackage{graphicx}
\usepackage{epsfig}
\usepackage{dsfont}
\usepackage{amsmath}
\usepackage{amssymb}
\usepackage{boldline,multirow}
\usepackage{siunitx}
\usepackage{booktabs}
\usepackage{soul}
\usepackage{cite}
\usepackage{cases}
\usepackage{hyperref}
\usepackage{bbding}
\newcommand{\ucite}[1]{\cite{#1}} 
\begin{document}
\title{Bounding Maps for Universal Lesion Detection}
\titlerunning{Bounding maps for universal lesion detection}
%
%


\newcommand*\samethanks[1][\value{footnote}]{\footnotemark[#1]}

\author{Han Li\inst{1,2} \and
Hu Han\inst{1,3}\Envelope \and
S. Kevin Zhou\inst{1,3}\Envelope}

\institute{
Medical Imaging, Robotics, Analytic Computing Laboratory/Engineering (MIRACLE), Key Laboratory of Intelligent Information Processing of Chinese Academy of Sciences (CAS),
Institute of Computing Technology, CAS, Beijing, China \email{\{hanhu, zhoushaohua\}@ict.ac.cn} \and
University of Chinese Academy of Sciences, Beijing, China \and
Peng Cheng Laboratory, Shenzhen, China}

%


\authorrunning{Han~Li et al.}

\maketitle              

\begin{abstract}
Universal Lesion Detection (ULD) in computed tomography plays an essential role in computer-aided diagnosis systems.  Many detection approaches achieve excellent results for ULD using possible bounding boxes (or anchors) as proposals. However, empirical evidence shows that using anchor-based proposals leads to a high false-positive (FP) rate.
In this paper, we propose a \textbf{box-to-map} method to represent a bounding box with three  soft continuous maps with bounds in $x$-, $y$- and $xy$- directions. The \textbf{bounding maps (BMs)} are used in two-stage anchor-based ULD frameworks to reduce the FP rate.
In the $1^{st}$ stage of the region proposal network, we replace the sharp binary ground-truth label of anchors with the corresponding  $xy$-direction BM
hence the positive anchors are now graded. In the $2^{nd}$ stage, we add a branch that takes our continuous BMs in $x$- and $y$- directions for extra supervision of detailed  \textbf{locations}. Our method, when embedded into three state-of-the-art two-stage anchor-based detection methods, brings a free detection  accuracy improvement (e.g., a 1.68\% to  3.85\% boost of sensitivity at 4 FPs) without extra inference time. \let\thefootnote\relax\footnotetext{This work is supported in part by the Natural Science Foundation of China (grants 61672496), Youth Innovation Promotion Association CAS (grant 2018135) and Alibaba Group through Alibaba Innovative Research Program.}
\end{abstract}
\section{Introduction}
Universal Lesion Detection (ULD)  in computed tomography (CT) images \ucite{zhang2019anchor_free,zhang2019lesion,zhang2020Agg_Fas,zlocha2019one-stage,tao2019improving,tang2019uldor,yan20183DCE,li2019mvp}, which aims to localize different types of lesions instead of identifying lesion types \ucite{liao2019evaluate,lin2019automated,wang2019volumetric,yan2019mulan,astaraki2019normal,tang2019nodulenet,shao2019attentive,liu20193dfpn,wang2018automated,zhu2018deepem,li2020high,liu20183d}, plays an essential role in computer-aided  diagnosis (CAD) systems. Recently, deep learning-based detection approaches achieve excellent results for ULD \cite{zhou2015medical}\cite{zhou2017deep} using possible bounding boxes (BBoxs) (or anchors) as proposals. However, empirical evidence shows that using anchor-based proposals leads to severe data imbalance (e.g., class and spatial imbalance) \cite{oksuz2020imbalance}, which leads to a high false-positive (FP) rate in ULD.  Therefore, there is an urgent need to reduce the FP proposals and improve the lesion detection performance.

Most existing ULD methods are mainly inspired by the successful deep models in object detection from natural images.  Tang et al. \cite{tang2019uldor} constructed a pseudo mask for each lesion region as the extra supervision information to adapt a  Mask-RCNN \cite{he2017maskrcnn} for ULD. Yan et al. \cite{yan20183DCE} proposed a 3D Context Enhanced (3DCE) R-CNN model based on the model \cite{deng2009imagenet} pre-trained from ImageNet for 3D context modeling. Li et al.  \cite{li2019mvp} proposed the so-called MVP-Net, which is a multi-view feature pyramid network (FPN) \cite{lin2017fpn} with position-aware attention to incorporate multi-view information for ULD. Han et al. \cite{8606226} leveraged cascaded multi-task learning to jointly optimize object detection and representation learning.

\begin{figure}[t]
\centering
\setlength{\abovecaptionskip}{0.cm}

\includegraphics[scale=0.36]{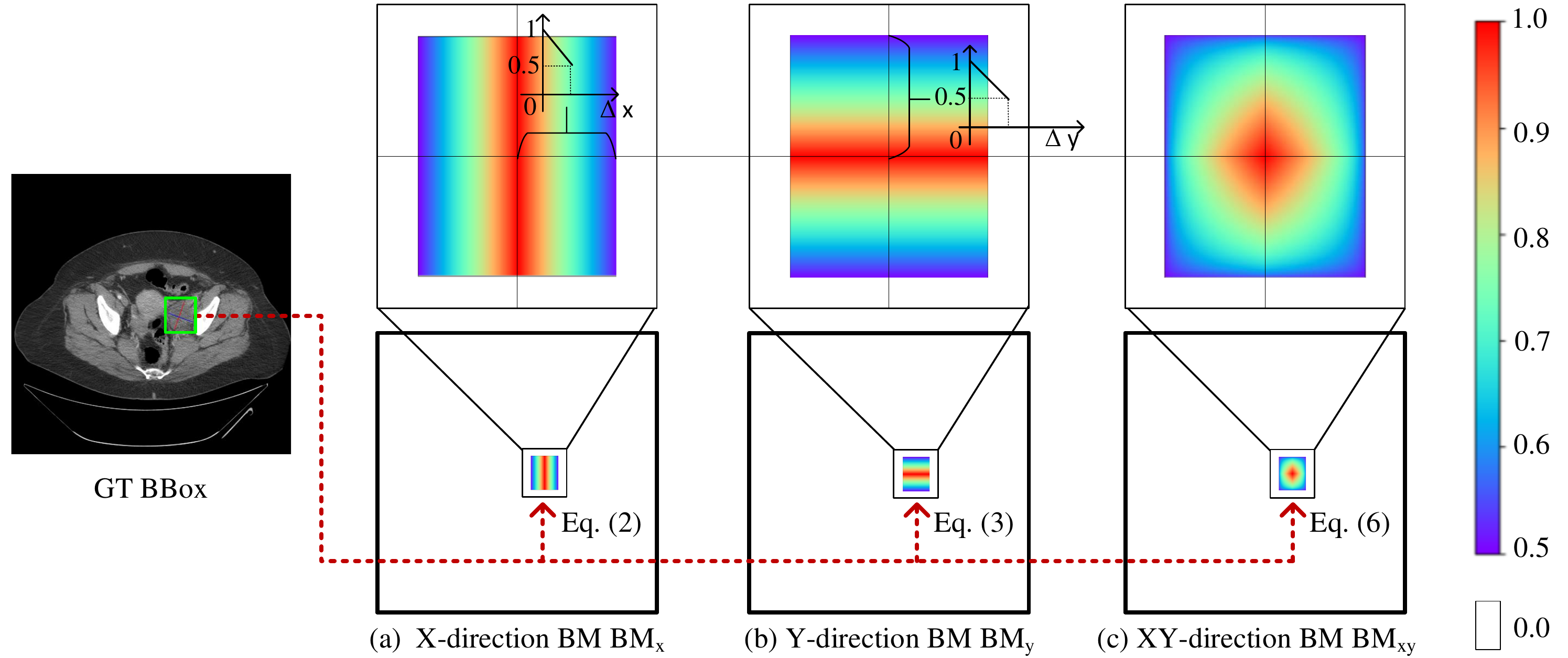}
\caption{The sharp GT BBox of an image is represented by three continuous 2D  bounding maps (BMs) in (along) different directions (axes): (a) $BM_x$, (b) $BM_y$, (c) $BM_{xy}$.}
\label{fig:fig1_generate_BM}

\end{figure}
All the above  approaches proposed for ULD  are designed based on a two-stage anchor-based framework, i.e., proposal generation followed by classification and regression like Faster R-CNN \cite{ren2015fasterrcnn}. They achieve good performance because:
i) The anchoring mechanism is a good reception field initialization for limited-data and limited-lesion-category datasets. ii) The two-stage mechanism is a coarse-to-fine mechanism for the CT lesion dataset that only contains two categories (`lesion' or not ), i.e., first finds lesion proposals and then removes the FP proposals.
However, such a framework has two main limitations for effective ULD:
(i) \emph{The imbalanced anchors in stage-1.} (e.g, class, spatial imbalance \cite{oksuz2020imbalance}). 
In the first stage, anchor-based methods first find out the positive (lesion) anchors and use them as the region of interest (ROI) proposals according to the intersection over union (IoU) between anchors and ground-truth (GT) BBoxs. Hence, the number of positive anchors is decided by the IoU threshold and the amount of GT BBoxs per image. Specifically, an anchor is considered positive if its IoU with a GT BBox is greater than the IoU threshold and negative otherwise.  This idea helps natural images to get enough positive anchors because they may have  a lot of GT BBoxs per image, but it isn't suitable for ULD. Most CT slices only have one or two GT lesion BBox(s), so the amount of positive anchors is rather limited. This limitation can cause severe data imbalance and influence the training convergence of the whole network. Using a lower IoU threshold is a simple way to get more positive anchors, but a lot of low-IoU anchors are labeled as positive can also lead to a high FP rate in ULD.
(ii) \emph{The insufficient supervision in stage-2.} 
In the second stage, each ROI proposal (selected anchor) from the first stage has one corresponding classification score to represent the possibility of containing lesions. The ROI proposals with high classification scores are chosen to obtain the final BBox prediction. ULD is a challenging task due to the similar appearances (e.g., intensity and texture) between lesions and other tissues; the non-lesion regions can also get very high scores. Hence, a single classification score can easily lead to FPs in ULD.

To address the anchor-imbalance problem, anchor-free methods \ucite{tian2019fcos,zhou2019objectsaspotints} solve detection in a per-pixel prediction manner and achieve success in natural images with sufficient data and object categories. But for lesion detection (lesion or not) with limited data, they lack needed precision. To overcome the supervision-insufficient problem, Mask R-CNN-like \cite{he2017maskrcnn} methods add a mask branch to introduce extra segmentation supervision and hence improve the detection performance. But it needs training segmentation masks that are costly to obtain.

\begin{figure}[t]
\centering
\setlength{\abovecaptionskip}{0.cm}

\includegraphics[scale=0.3]{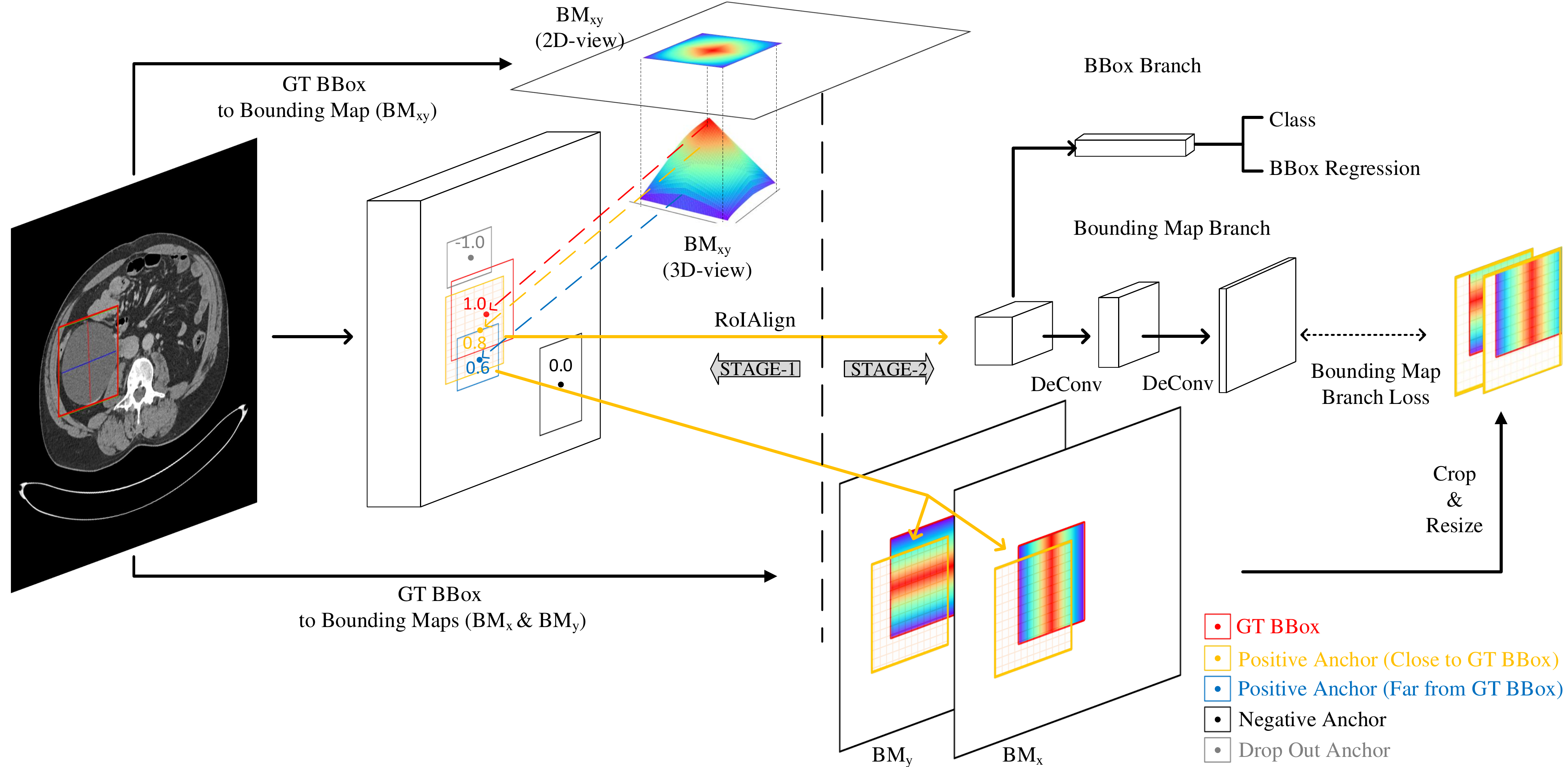}
\caption{The network architecture of the proposed ULD approach, in which, three proposed BMs in $x$-, $y$- and $xy$- directions are used in its two stages.}
\label{fig:fig2_network_architecture}

\end{figure}

In this paper, we present a continuous bounding map (BM) representation to enable the per-pixel prediction in the 1st stage and introduce extra-supervision in the 2nd stage of any anchor-based detection method.  Our \textbf{first contribution} is a new box-to-map representation, which represents a BBox by three 2D bounding maps (BMs) in (along) three different directions (axes): $x$-direction ($BM_x$), $y$-direction ($BM_y$), and $xy$-direction ($BM_{xy}$), as shown in Fig. \ref{fig:fig1_generate_BM}.
The pixel values in $BM_x$ and $BM_y$ decrease from the centerline to the BBox borders in x and y directions respectively with a linear fashion, while the pixel values in $BM_{xy}$ decrease from both two directions. Compared with a sharp binary representation (e.g., binary anchors label in RPN, binary segmentation mask in Mask R-CNN \cite{he2017maskrcnn}), such a soft continuous map can provide a more detailed representation of \textbf{location}. This (i.e., per-pixel \& continuous) promotes the network to learn more contextual information \cite{zhou2019objectsaspotints}, thereby reducing the FPs. Our \textbf{second contribution} is to expand the capability of a two-stage anchor-based detection method using our BM representation in a light way.  First, we use $BM_{xy}$ as the GT of a positive anchor in the first stage as in Fig. \ref{fig:fig2_network_architecture} and choose a proper IoU threshold to deal with the anchor imbalance problem. Second, we add one additional branch called BM branch paralleled with the BBox branch \cite{ren2015fasterrcnn} in the second stage as in Fig. \ref{fig:fig3_BM_branch}. The BM branch introduces extra supervision of detailed location to the whole network in a pixel-wise manner and thus decreases the FP rate in ULD. We conduct extensive experiments on the DeepLesion Dataset \cite{yan18deeplesion} with four state-of-the-art ULD methods to validate the effectiveness of our method.


\section{Method}
As shown in Fig. \ref{fig:fig2_network_architecture}, we utilize BMs to reduce the ULD FP rate by replacing the original positive anchor class labels in stage-1 and adding a BM branch to introduce extra pixel-wise location supervision in stage-2. Section \ref{sec:bm} details the BM representation and Section \ref{sec:bmanchor} defines the anchor labels for RPN training based on our BMs. Section \ref{sec:bmbranch} explains the newly introduced BM branch.


\subsection{Bounding maps} \label{sec:bm}

Motivated by \cite{zhou2019objectsaspotints}, the BMs are formed in all-zero maps by only changing the value of pixels located within the BBox(s) as  in Fig. \ref{fig:fig1_generate_BM}. Let $(x^{(i)}_1,y^{(i)}_1,x^{(i)}_2,y^{(i)}_2)$ be the $i^{th}$ lesion GT BBox of one CT image $I_{ct}\in \mathcal{R}^ {W\times H}$, the set of coordinates within $i^{th}$ BBox can be denoted as:
\begin{equation}
  S^{(i)}_{BBox}=\{(x,y){|}x^{(i)}_1 \leq x \le x^{(i)}_2 ~\&~ y^{(i)}_1 \leq y \le y^{(i)}_2 \},
\end{equation}
and the center point of this BBox lies at $(x^{(i)}_{ctr},y^{(i)}_{ctr})=(\frac{x^{(i)}_1+x^{(i)}_2}{2},\frac{y^{(i)}_1+y^{(i)}_2}{2})$.

Within each BBox $S^{(i)}_{BBox}$, the pixel values in $BM^{(i)}_x \in \mathcal{R}^ {W\times H}$ and $BM^{(i)}_y \in \mathcal{R}^ {W\times H}$ decrease from 1 (center line) to 0.5 (border)  in a  linear fashion:
\begin{equation}
BM^{(i)}_x(x,y)=\left\{
\begin{array}{lp{8mm}<{\centering}l}
0& & {(x,y) \notin S^{(i)}_{BBox}}\\
1-k^{(i)}_x\left|x^{(i)}-x^{(i)}_{ctr}\right| & &{(x,y) \in S^{(i)}_{BBox}}
\end{array} \right. ,
\end{equation}

\begin{equation}
BM^{(i)}_y(x,y)=\left\{
\begin{array}{lp{8mm}<{\centering}l}
0& & {(x,y) \notin S^{(i)}_{BBox}}\\
1-k^{(i)}_y\left|y^{(i)}-y^{(i)}_{ctr}\right| & &{(x,y) \in S^{(i)}_{BBox}}
\end{array} \right. ,
\end{equation}
where $k^{(i)}$ is the slope of linear function in $x$-direction or $y$-direction, which  is calculated according to the GT BBox's width ($x^{(i)}_2-x^{(i)}_{1}$) or height ($y^{(i)}_2-y^{(i)}_{1}$):
\begin{equation}
  k^{(i)}_x= \frac{1}{ x^{(i)}_2-x^{(i)}_{1}}, ~~k^{(i)}_y= \frac{1}{ y^{(i)}_2-y^{(i)}_{1}}.
\end{equation}
We take the sum of all the $BM^{(i)}_x$s and  $BM^{(i)}_y$s to obtain the total $BM_x \in \mathcal{R}^ {W\times H}$ and $BM_y \in \mathcal{R}^ {W\times H}$ of one input image, respectively.

 \begin{equation}
   BM_x=\min \Big [ \sum\limits_{i=1}^I BM^{(i)}_x,1 \Big ], BM_y=\min  \Big [\sum\limits_{i=1}^I BM^{(i)}_y, 1  \Big ],
  \end{equation}
where $I$ is the number of GT BBox(s) of one CT image.
Then the $xy$-direction BM $BM_{xy}\in \mathcal{R}^ {W\times H}$ can be generated by calculating the square root of the product between $BM_x$ and $BM_y$:
\begin{equation}
  BM_{xy}=\sqrt[2]{BM_{x}\odot BM_{y} },
\end{equation}
where $\odot$ denotes the element-wise multiplication.

By introducing the above BMs, we expect they can promote network training and reduce FPs. Because the proposed BMs offer a soft continuous map about the lesion other than a sharp binary mask, which can convey more contextual information about the lesion, not only its location but also guidance of confidence. These properties are favourable for object detection task with irregular shapes and limited-amount GT BBox(s) like ULD.

\subsection{Anchor label in RPN} \label{sec:bmanchor}

In the original two-stage anchor-based detection frameworks, RPN is trained to produce  object bounds and objectness classification scores $\hat{C}_{ct} \in \mathcal{R}^ {\frac{W}{R}\times \frac{H}{R} \times 1}$ at each position (or anchors' centerpoint), where $R$ is the output stride. During training, all the anchors are first divided into three categories of positive (lesion), negative and drop-out anchors based on their IoUs with the GT BBoxs. Then the GT labels of positive, negative  and drop-out anchors are set as 1, 0, -1 respectively and only the positive and negative anchors are used for loss calculation in RPN training.

In our proposed method, we still use 0 and -1 as  the GT class labels of negative and drop-out anchors, but we set the class label of positive anchors as their corresponding value in $BM_{xy}\in \mathcal{R}^ {W\times H} $. For size consistency, we first resize $BM_{xy}\in \mathcal{R}^ {W\times H} $ to $BM^r_{xy}\in \mathcal{R}^  {\frac{W}{R}\times \frac{H}{R} \times 1} $ to match the size of $\hat{C}_{ct} \in \mathcal{R}^ {\frac{W}{R}\times \frac{H}{R} \times 1}$. 
Therefore, the GT label of anchor $C_{anc}$ is given as:

  \begin{equation}
    C_{anc}(x,y,IoU_{anc})=\left\{
  \begin{array}{lp{8mm}<{\centering}c}
    0&&{IoU_{anc} \leq IoU_{min}}\\
  -1& & {IoU_{min}<IoU_{anc}<IoU_{max}}\\

  BM^{r}_{xy}[x,y] & &{IoU_{anc}\geq IoU_{max}}
\end{array} \right.,
  \end{equation}
where $(x,y)$ is the centerpoint coordinates of an anchor in $\hat{C}_{ct} \in \mathcal{R}^ {\frac{W}{R}\times \frac{H}{R} \times 1}$ and $IoU_{anc}$ denotes the IoU between the anchor and GT BBox.

\textbf{Anchor classification loss function:} For each anchor, the original RPN loss is the sum of anchor classification loss and BBox regression loss. However, the amount of GT BBox in one CT slice is usually more limited than one natural image. Hence a proper RPN IoU threshold is hard to find in ULD task: a higher IoU threshold can cause imbalanced anchors problem while a lower IoU threshold which causes too many low-IoU anchors' GT label are set as $1$ can lead to a high FP rate. Therefore, we replace the original anchor classification loss with our proposed anchor classification loss:

  \begin{equation}
  \mathcal{L}_{anc}=\left\{
\begin{array}{lp{1mm}<{\centering}c}
  \mathcal{L}_2(\hat{C}_{anc},C_{anc})&&{IoU_{anc} \leq IoU_{min}~\&IoU_{anc}\geq IoU_{max}}\\
0& & {IoU_{min}<IoU_{anc}<IoU_{max}}
\end{array} \right.,
  \end{equation}
  where $\mathcal{L}_2$ is the norm-2 loss, $IoU_{min}$ and $IoU_{max}$ denote the negative and positive anchor thresholds, respectively.

  \begin{figure}[t]
  \centering
  \setlength{\abovecaptionskip}{-0.05cm}

   \includegraphics[scale=0.23]{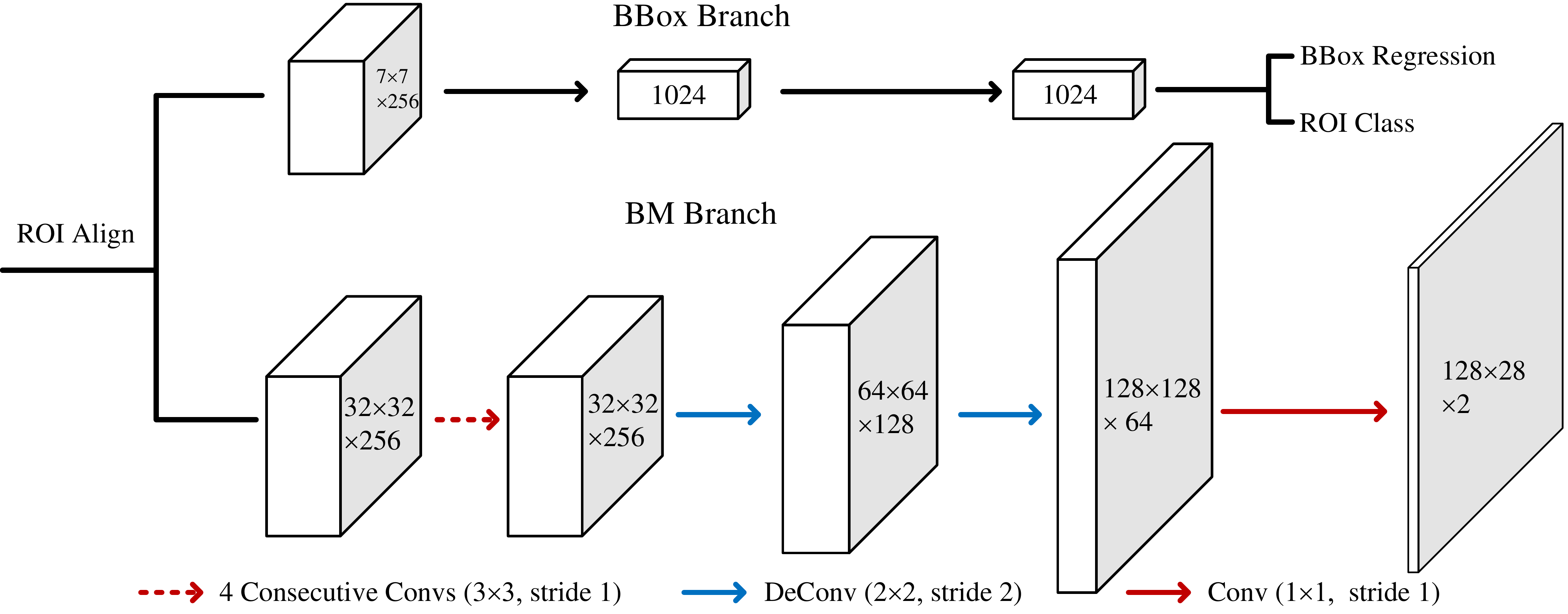}
  \caption{BM branch is in parallel with BBox branch and applied separately to each ROI.} 
  \label{fig:fig3_BM_branch}

  \end{figure}

 \subsection{Bounding map branch} \label{sec:bmbranch}
As shown in Fig. \ref{fig:fig3_BM_branch}, the BM branch is similar to the mask branch in Mask R-CNN \cite{he2017maskrcnn}. It is paralleled with the BBox branch and applied separately to each ROI. The branch consists of  four $3\times 3$ convolution, two $2 \times 2$ deconvolution and one $1 \times 1$ convolution layers. It takes ROI proposal feature map $F_{ROI} \in \mathcal{R}^ {32\times 32 \times 256}$ as input and aims to obtain the $BM_x$ and $BM_y$ proposals, denoted by $BM^{ROI}_x \in \mathcal{R}^ {128\times 128 \times 1}$ and $BM^{ROI}_y\in \mathcal{R}^ {128\times 128 \times 1}$, respectively:
\begin{equation}
   [ \hat{BM}^{ROI}_x , \hat{BM}^{ROI}_y ] = H_{BM}(F_{ROI}),
\end{equation}
where $H_{BM}$ is the function of BM branch. 

\textbf{BM branch loss function:}
For each ROI, $BM_x$ and $BM_y$ are first cropped based on the ROI BBox and resized to the size of the BM branch output to obtain  $BM^{ROI}_x \in \mathcal{R}^ {128\times 128 \times 1}$ and $BM^{ROI}_y\in \mathcal{R}^ {128\times 128 \times 1}$. Then we concatenate the two BMs into a multi-channel map and use it as the ground-truth for our BM branch. Therefore, the loss function of BM branch for each ROI can be defined as a norm-2 loss:

\begin{equation}
\mathcal{L}_{BM}=\mathcal{L}_2([\hat{BM}^{ROI}_x,\hat{BM}^{ROI}_y],[BM^{ROI}_x,BM^{ROI}_y]).
\end{equation}

\textbf{Full loss function:}
The full loss function of our method is given as:
\begin{equation}
\mathcal{L}_{full}=\frac{1}{M}\sum\limits_{m=1}^M(\mathcal{L}^{(m)}_{reg}+\mathcal{L}^{(m)}_{anc})+\frac{1}{N}\sum\limits_{n=1}^N(\mathcal{L}^{(n)}_{B}+\mathcal{L}^{(n)}_{BM}),
\end{equation}
where $\mathcal{L}^{(m)}_{reg}$ and $\mathcal{L}^{(n)}_{B}$ are the original box regression loss (in RPN) of the $m^{th}$ training (negative $\&$ positive) anchor and BBox branch loss  of $n^{th}$ positive ROI in Faster R-CNN \cite{ren2015fasterrcnn}. $\mathcal{L}^{(m)}_{anc}$ and $\mathcal{L}^{(n)}_{BM}$ denote our anchor classification loss and BM branch loss for $m^{th}$ training anchors and $n^{th}$ positive ROI. $M$ and $N$ are the number of training anchors and positive ROIs, respectively.
\begin{table}[t]
\centering
\caption{Sensitivity (\%) at various FPPI on the testing set of DeepLesion \cite{yan18deeplesion}.} \label{results}
\newsavebox{\tablebox}
\begin{lrbox}{\tablebox}
\begin{tabular}{p{55mm}p{25mm}<{\centering}p{25mm}<{\centering}p{25mm}<{\centering}p{25mm}<{\centering}p{25mm}<{\centering}p{25mm}<{\centering}}
\toprule
\textbf{FPPI}&\textbf{$@0.5$}&\textbf{$@1$}&\textbf{$@2$}&\textbf{$@3$}&\textbf{$@4$}\\
\midrule
Faster R-CNN \cite{ren2015fasterrcnn}&57.17 & 68.82 &74.97 &78.48&82.43\\
Faster R-CNN w/ Ours&63.96 (\textbf{6.79}$\uparrow$) & 74.43 (\textbf{5.61}$\uparrow$) &79.80 (\textbf{4.83}$\uparrow$) &82.55 (\textbf{4.07}$\uparrow$)&86.28  (\textbf{3.85}$\uparrow$) \\
\midrule
3DCE (9 slices) \cite{yan20183DCE}  &59.32 &70.68&79.09&-&84.34\\
3DCE (9 slices) w/ Ours &64.38 (5.06$\uparrow$)&75.55 (4.87$\uparrow$) &82.74  (3.65$\uparrow$)& 83.77 (-)& 87.78 (3.44$\uparrow$)\\
\midrule
3DCE (27 slices) \cite{yan20183DCE}  &62.48&73.37&80.70 &-&85.65 \\
3DCE (27 slices) w/ Ours &66.75 (4.27$\uparrow$)&76.71 (3.34$\uparrow$) &83.75  (3.05$\uparrow$)& 86.01 (-)& 88.59 (2.94$\uparrow$) \\
\midrule
FPN-3DCE (9 slices) \cite{lin2017fpn}& 64.25&74.41&81.90&85.02&87.21\\
FPN-3DCE (9 slices) w/ Ours&69.09 (4.84$\uparrow$)&78.02 (3.61$\uparrow$) &85.35  (3.45$\uparrow$)& 88.59 (3.57$\uparrow$)& 90.49 (3.28$\uparrow$)\\

\midrule
MVP-Net (3 slices) \cite{li2019mvp}&70.01&78.77&84.71&87.58&89.03\\
MVP-Net (3 slices) w/ Ours&\textbf{73.32} (3.31$\uparrow$)&\textbf{81.24} (2.47$\uparrow$) &\textbf{86.75} (2.04$\uparrow$)& \textbf{89.54} (1.96$\uparrow$)& \textbf{90.71} (1.68$\uparrow$)\\
\midrule
FCOS (anchor-free) \cite{tian2019fcos}&37.78&54.84&64.12&69.41&77.84\\
\midrule
Objects as points (anchor-free) \cite{zhou2019objectsaspotints}&34.87&43.58&52.41&59.13&64.01\\
\bottomrule
\end{tabular}

\end{lrbox}
\scalebox{0.65}{\usebox{\tablebox}}
\end{table}
\section{Experiments}
\subsection{Dataset and setting}
We conduct experiments using the DeepLesion dataset \cite{yan18deeplesion}.  The dataset  is a large-scale CT dataset with 32,735 lesions on 32,120 axial slices
from 10,594 CT studies of 4,427 unique patients. Different from existing datasets that typically focus on one type of lesion, DeepLesion contains a variety of lesions with a variety of diameters ranges (from 0.21 to 342.5mm). We rescale the 12-bit CT intensity range to [0,255] with different window ranges proposed in different frameworks. Every CT slice is resized to $800\times800$, and the slice intervals are interpolated to 2mm.\footnote{We use a CUDA toolkit in \cite{huang20193d} to speed up this process.} We conducted experiments on the official training ($70\%$), validation ($15\%$), testing ($15\%$) sets. The number of FPs per image (FPPI) is used as the evaluation metric, and we mainly compare the sensitivity at 4 FPPI for WRITING briefness,  just as in \cite{li2019mvp}.

We only use the horizontal flip as the training data augmentation and train them with stochastic gradient descent (SGD) for $15$ epochs. The base learning rate is set as $0.002$, and decreased by a factor of $10$ after the $12^{th}$ and $14^{th}$ epoch. The models with our method utilize a lower positive anchor IoU threshold of 0.5, and the other network settings are the same as the corresponding original models.

\subsection{Detection performance}

We perform experiment with three state-of-the-art two-stage anchor-based detection methods to evaluate the effectiveness of our approach. We also use two state-of-the-art anchor-free natural image detection methods for comparison.
\begin{itemize}
    \item \textbf{3DCE.} The  3D context enhanced region-based CNN (3DCE) \cite{yan20183DCE}  is trained with 9 or 27 CT slices to form the 9-slice or 27-slice 3DCE.
    \item \textbf{FPN-3DCE.} The 3DCE \cite{yan20183DCE} is re-implemented with the FPN backbone \cite{lin2017fpn} and trained with 9 CT slices to form the 9-slice  FPN-3DCE. The other network setting is consistent with the baseline 3DCE.
    \item \textbf{MVP-Net.} The multi-view FPN with position-aware attention network (MVP-Net) \cite{li2019mvp} is trained with 3 CT slices to form the 3-slice MVP-Net.
    \item \textbf{Faster R-CNN. } Similar to MVP-Net \cite{li2019mvp}, we rescale an original 12-bit CT image with window ranges of  [50,449], [-505,1980] and [446,1960] to generate three rescaled  CT images. Then we concatenate the three rescaled  CT images into three channels to train a  Faster R-CNN. The other network settings are the same as the baseline MVP-Net.
    \item\textbf{FCOS \& Objects as points.} The experiment settings for the anchor-free methods, namely Fully Convolutional One-Stage Object Detection (FCOS) \cite{tian2019fcos} and objects as points \cite{zhou2019objectsaspotints}, are the same as the baseline Faster R-CNN.
\end{itemize}

\begin{figure}[t]
\centering
\setlength{\abovecaptionskip}{0.cm}

\includegraphics[scale=0.26]{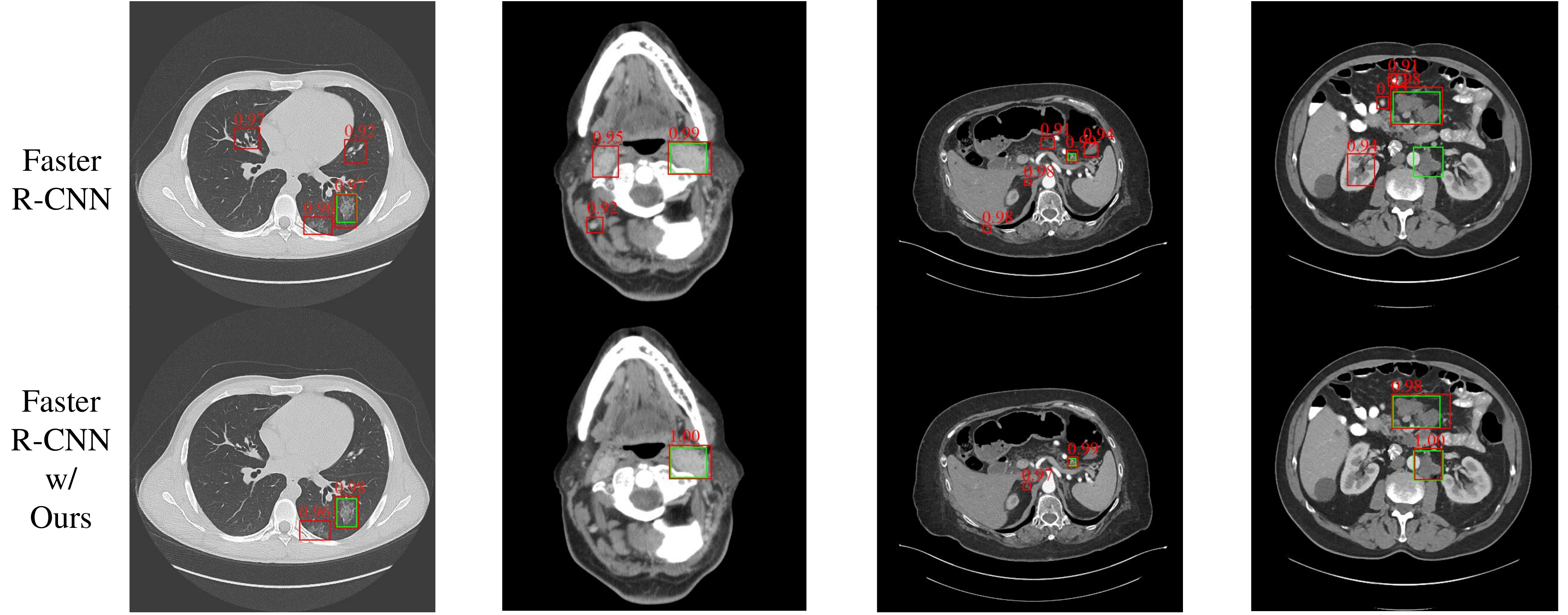}
\caption{ High-classification-score results (above 0.9) of Faster R-CNN with or without our method on a test image. Green and red boxes corresponded to GT BBox and predicted BBox, respectively. The classification scores are marked in the images.}
\label{fig:fig4_vis_results}

\end{figure}

As shown in Table  \ref{results},  our method brings promising detection performance improvements for all baselines uniformly at different FPPIs. The improvement of Faster R-CNN \cite{ren2015fasterrcnn}, 9-slice 3DCE, 27-slice 3DCE and  9-slice FPN-3DCE are more pronounced than that of MVP-Net. This is because the MVP-Net is designed for reducing the FP rate in UDL and has achieved relatively high performance. Also, the anchor-free methods yields unsatisfactory results, and we think the main reason is that they completely discard the anchor and two-stage mechanism. Fig. \ref{fig:fig4_vis_results} presents a case to illustrate the effectiveness of our method in improving the performance of Faster-R-CNN.

\subsection{Ablation study}
We provide an ablation study about the two key components of the proposed approach, e.g., with vs. without using $BM_{xy}$  in stage-1 and with vs. without using BM branch ($BM_{x} ~\&~ BM_{y}$) in stage-2. We also perform a study to compare the efficiency between linear BMs and Gaussian BMs. As shown in Table \ref{ablation_study},  using $BM_{xy}$ as the  class label for positive anchors, we obtain a 2.27\% improvement over the Faster R-CNN \cite{ren2015fasterrcnn} baseline. Further adding a BM branch for introducing extra pixel-wise supervision  accounts for another 1.14\% improvement.  Using both $BM_{xy}$ and BM branch gives the best performance. Taking Gaussian BM instead linear BM does not bring improvement. The use of our method causes a minor influence to the  inference time measured on a Titan XP GPU.
\begin{table}[t]
\centering
\caption{Ablation study of our method at various FPs per image (FPPI).} 
\label{ablation_study}
\begin{lrbox}{\tablebox}

\begin{tabular}{p{30mm}<{\centering}p{20mm}<{\centering}p{20mm}<{\centering}p{25mm}<{\centering}|p{20mm}<{\centering}p{20mm}<{\centering}|p{30mm}<{\centering}}
\toprule
Faster R-CNN \cite{ren2015fasterrcnn}&$BM_{xy}$&$BM_x \& BM_y$&Gaussian $BM$ &$FPPI=2$&$FPPI=4$&Inference (s/img)\\
$\checkmark$&&&&74.97&78.48&0.3912\\
$\checkmark$&$\checkmark$&&&77.47&80.69&0.3946\\
$\checkmark$&$\checkmark$&$\checkmark$&&\textbf{79.80}&\textbf{82.55}&0.3946\\
$\checkmark$&$\checkmark$&$\checkmark$&$\checkmark$&78.44&{82.37}&0.4004\\

\bottomrule
\end{tabular}

\end{lrbox}
\scalebox{0.7}{\usebox{\tablebox}}
\end{table}

\section{Conclusion}
In this paper, we study how to overcome the two limitations of two-stage anchor-based ULD methods: the imbalanced anchors in the first stage and the insufficient supervision information in the second stage. We first propose BMs to represent a BBox in three different directions and then use them to replace the original binary GT labels of positive anchors in stage-1 introduce additional supervision through a new  BM branch in stage-2.
We conduct experiments based on several state-of-the-art baselines on the DeepLesion dataset, and the results show that the performances of all the baselines are boosted with our method.


\bibliographystyle{splncs04}
\bibliography{egbib}

\begin{thebibliography}{10}

\bibitem{zhang2019anchor_free}
N.~Zhang, D.~Wang, X.~Sun, P.~Zhang, C.~Zhang, Y.~Cao, and B.~Liu.
\newblock 3d anchor-free lesion detector on computed tomography scans.
\newblock {\em arXiv:1908.11324}, 2019.

\bibitem{zhang2019lesion}
Z.~Zhang, Y.~Zhou, W.~Shen, E.~Fishman, and A.~Yuille.
\newblock Lesion detection by efficiently bridging 3d context.
\newblock In {\em MLMI Workshop}, pages 470--478. Springer, 2019.

\bibitem{zhang2020Agg_Fas}
N.~Zhang, Y.~Cao, B.~Liu, and Y.~Luo.
\newblock 3d aggregated faster {R-CNN} for general lesion detection.
\newblock {\em arXiv:2001.11071}, 2020.

\bibitem{zlocha2019one-stage}
M.~Zlocha, Q.~Dou, and B.~Glocker.
\newblock Improving retinanet for ct lesion detection with dense masks from
  weak recist labels.
\newblock In {\em MICCAI}, pages 402--410. Springer, 2019.

\bibitem{tao2019improving}
Q.~Tao, Z.~Ge, J.~Cai, J.~Yin, and S.~See.
\newblock Improving deep lesion detection using 3d contextual and spatial
  attention.
\newblock In {\em MICCAI}, pages 185--193. Springer, 2019.

\bibitem{tang2019uldor}
Y.~Tang, K.~Yan, Y.~Tang, J.~Liu, J.~Xiao, and R.~M Summers.
\newblock Uldor: a universal lesion detector for ct scans with pseudo masks and
  hard negative example mining.
\newblock In {\em IEEE ISBI}, pages 833--836, 2019.

\bibitem{yan20183DCE}
K.~Yan, M.~Bagheri, and R.~M Summers.
\newblock 3d context enhanced region-based convolutional neural network for
  end-to-end lesion detection.
\newblock In {\em MICCAI}, pages 511--519. Springer, 2018.

\bibitem{li2019mvp}
Z.~Li, S.~Zhang, J.~Zhang, K.~Huang, Y.~Wang, and Y.~Yu.
\newblock Mvp-net: Multi-view fpn with position-aware attention for deep
  universal lesion detection.
\newblock In {\em MICCAI}, pages 13--21. Springer, 2019.

\bibitem{liao2019evaluate}
F.~Liao, M.~Liang, Z.~Li, X.~Hu, and S.~Song.
\newblock Evaluate the malignancy of pulmonary nodules using the 3-d deep leaky
  noisy-or network.
\newblock {\em IEEE Trans. Neural Netw. Learn. Syst}, 30(11):3484--3495, 2019.

\bibitem{lin2019automated}
Y.~Lin, J.~Su, X.~Wang, X.~Li, J.~Liu, K.~Cheng, and X.~Yang.
\newblock Automated pulmonary embolism detection from ctpa images using an
  end-to-end convolutional neural network.
\newblock In {\em MICCAI}, pages 280--288. Springer, 2019.

\bibitem{wang2019volumetric}
X.~Wang, S.~Han, Y.~Chen, D.~Gao, and N.~Vasconcelos.
\newblock Volumetric attention for 3d medical image segmentation and detection.
\newblock In {\em MICCAI}, pages 175--184. Springer, 2019.

\bibitem{yan2019mulan}
K.~Yan, Y.~Tang, Y.~Peng, V.~Sandfort, M.~Bagheri, Z.~Lu, and R.~M Summers.
\newblock Mulan: Multitask universal lesion analysis network for joint lesion
  detection, tagging, and segmentation.
\newblock In {\em MICCAI}, pages 194--202. Springer, 2019.

\bibitem{astaraki2019normal}
M.~Astaraki, I.~Toma-Dasu, {\"O}.~Smedby, and C.~Wang.
\newblock Normal appearance autoencoder for lung cancer detection and
  segmentation.
\newblock In {\em MICCAI}, pages 249--256. Springer, 2019.

\bibitem{tang2019nodulenet}
C.~Tang, H.and~Zhang and X.~Xie.
\newblock Nodulenet: Decoupled false positive reduction for pulmonary nodule
  detection and segmentation.
\newblock In {\em MICCAI}, pages 266--274. Springer, 2019.

\bibitem{shao2019attentive}
Q.~Shao, L.~Gong, K.~Ma, H.~Liu, and Y.~Zheng.
\newblock Attentive ct lesion detection using deep pyramid inference with
  multi-scale booster.
\newblock In {\em MICCAI}, pages 301--309. Springer, 2019.

\bibitem{liu20193dfpn}
J.~Liu, L.~Cao, O.~Akin, and Y.~Tian.
\newblock 3dfpn-hs: 3d feature pyramid network based high sensitivity and
  specificity pulmonary nodule detection.
\newblock In {\em MICCAI}, pages 513--521. Springer, 2019.

\bibitem{wang2018automated}
B.~Wang, G.~Qi, S.~Tang, L.~Zhang, L.~Deng, and Y.~Zhang.
\newblock Automated pulmonary nodule detection: High sensitivity with few
  candidates.
\newblock In {\em MICCAI}, pages 759--767. Springer, 2018.

\bibitem{zhu2018deepem}
W.~Zhu, Y.~S Vang, Y.~Huang, and X.~Xie.
\newblock Deepem: Deep 3d convnets with em for weakly supervised pulmonary
  nodule detection.
\newblock In {\em MICCAI}, pages 812--820. Springer, 2018.

\bibitem{li2020high}
H.~Li, H.~Han, Z.~Li, L.~Wang, Z.~Wu, J.~Lu, and S.~Kevin Zhou.
\newblock High-resolution chest x-ray bone suppression using unpaired ct
  structural priors.
\newblock {\em IEEE Trans. Med. Imag.}, 2020.

\bibitem{liu20183d}
S.~Liu, D.~Xu, S~Kevin Zhou, O.~Pauly, S.~G., T.~Mertelmeier, J.~Wicklein,
  A.~Jerebko, W.~Cai, and D.~Comaniciu.
\newblock 3d anisotropic hybrid network: Transferring convolutional features
  from 2d images to 3d anisotropic volumes.
\newblock In {\em MICCAI}, pages 851--858. Springer, 2018.

\bibitem{zhou2015medical}
S.~Kevin Zhou.
\newblock {\em Medical image recognition, segmentation and parsing: machine
  learning and multiple object approaches}.
\newblock Academic Press, 2015.

\bibitem{zhou2017deep}
S.~Kevin Zhou, H.~Greenspan, and D.~Shen.
\newblock {\em Deep learning for medical image analysis}.
\newblock Academic Press, 2017.

\bibitem{oksuz2020imbalance}
K.~Oksuz, B.~C. Cam, S.~Kalkan, and E.~Akbas.
\newblock Imbalance problems in object detection: A review.
\newblock {\em Trans. Pattern Anal. Mach. Intell.}, 2020.

\bibitem{he2017maskrcnn}
K.~He, G.~Gkioxari, P.~Doll{\'a}r, and R.~Girshick.
\newblock {MASK R-CNN}.
\newblock In {\em IEEE ICCV}, pages 2961--2969, 2017.

\bibitem{deng2009imagenet}
J.~Deng, W.~Dong, R.~Socher, L.~Li, K.~Li, and L.~Fei-Fei.
\newblock Imagenet: A large-scale hierarchical image database.
\newblock In {\em IEEE CVPR}, pages 248--255, 2009.

\bibitem{lin2017fpn}
T.~Lin, P.~Doll{\'a}r, R.~Girshick, K.~He, B.~Hariharan, and S.~Belongie.
\newblock Feature pyramid networks for object detection.
\newblock In {\em IEEE CVPR}, pages 2117--2125, 2017.

\bibitem{8606226}
H.~{Han}, J.~{Li}, A.~K. {Jain}, S.~{Shan}, and X.~{Chen}.
\newblock Tattoo image search at scale: Joint detection and compact
  representation learning.
\newblock {\em IEEE Trans. Pattern Anal. Mach. Intell}, 41(10):2333--2348,
  2019.

\bibitem{ren2015fasterrcnn}
S.~Ren, K.~He, R.~Girshick, and J.~Sun.
\newblock {Faster R-CNN}: Towards real-time object detection with region
  proposal networks.
\newblock In {\em NIPS}, pages 91--99, 2015.

\bibitem{tian2019fcos}
Z.~Tian, C.~Shen, H.~Chen, and T.~He.
\newblock Fcos: Fully convolutional one-stage object detection.
\newblock In {\em IEEE ICCV}, pages 9627--9636, 2019.

\bibitem{zhou2019objectsaspotints}
X.~Zhou, D.~Wang, and P.~Kr{\"a}henb{\"u}hl.
\newblock Objects as points.
\newblock {\em arXiv:1904.07850}, 2019.

\bibitem{yan18deeplesion}
K.~Yan, X.~Wang, L.~Lu, L.~Zhang, A.~P Harrison, M.~Bagheri, and R.~M Summers.
\newblock Deep lesion graphs in the wild: relationship learning and
  organization of significant radiology image findings in a diverse large-scale
  lesion database.
\newblock In {\em IEEE CVPR}, pages 9261--9270, 2018.

\bibitem{huang20193d}
C.~Huang, H.~Han, Q.~Yao, S.~Zhu, and S~Kevin Zhou.
\newblock 3{D} ${U^{2}}$-net: A 3d universal u-net for multi-domain medical
  image segmentation.
\newblock In {\em MICCAI}, pages 291--299, 2019.

\end{thebibliography}

\end{document}